\documentclass[letterpaper]{article}

\usepackage[preprint]{aaai2027}
\usepackage[hyphens]{url}
\usepackage{graphicx}
\usepackage{tikz}
\usepackage{pifont}
\usetikzlibrary{positioning,fit,backgrounds,arrows.meta}
\usepackage{natbib}
\usepackage{caption}
\usepackage{booktabs}
\usepackage{amsmath}
\usepackage{amssymb}
\usepackage{xcolor}
\newcommand{\kSamples}{5}
\newcommand{\samplingTemp}{0.7}
\newcommand{\samplingSeed}{42}

\newcommand{\gptNoProfile}{0.461}
\newcommand{\gptRAG}{0.623}
\newcommand{\gptSimA}{0.722}
\newcommand{\gptFlat}{0.764}
\newcommand{\gptSim}{0.828}
\newcommand{\gptSimG}{0.856}

\newcommand{\llamaNoProfile}{0.353}
\newcommand{\llamaRAG}{0.500}
\newcommand{\llamaSimA}{0.501}
\newcommand{\llamaFlat}{0.605}
\newcommand{\llamaSim}{0.710}
\newcommand{\llamaSimG}{0.804}

\newcommand{\claudeNoProfile}{0.535}
\newcommand{\claudeRAG}{0.810}
\newcommand{\claudeSimA}{0.887}
\newcommand{\claudeFlat}{0.924}
\newcommand{\claudeSim}{0.954}
\newcommand{\claudeSimG}{0.956}

\newcommand{\ceilClaudeSim}{35}
\newcommand{\ceilClaudeSimG}{34}
\newcommand{\ceilClaudeFlat}{32}
\newcommand{\ceilClaudeRAG}{17}
\newcommand{\ceilGPTsim}{8}
\newcommand{\ceilLlamaSim}{2}
\newcommand{\groundingLlamaP}{0.0004}

\newcommand{\simVsRagGPT}{+0.205}
\newcommand{\simVsRagGPTci}{[0.139, 0.275]}
\newcommand{\simVsRagLlama}{+0.210}
\newcommand{\simVsRagClaude}{+0.144}
\newcommand{\simVsRagClaudeCI}{[0.080, 0.214]}
\newcommand{\simVsRagLlamaCI}{[0.137, 0.287]}

\newcommand{\profileVsNoneGPT}{+0.367}
\newcommand{\profileVsNoneLlama}{+0.357}
\newcommand{\profileVsNoneClaude}{+0.418}
\newcommand{\simVsFlatGPT}{+0.064}
\newcommand{\simVsFlatGPTp}{0.010}
\newcommand{\simVsFlatLlama}{+0.105}
\newcommand{\simVsFlatClaude}{+0.030}
\newcommand{\simVsFlatClaudeP}{0.256}

\newcommand{\arbitrationGPT}{+0.106}
\newcommand{\arbitrationLlama}{+0.209}
\newcommand{\arbitrationClaude}{+0.066}

\newcommand{\groundingGPT}{+0.029}
\newcommand{\groundingGPTci}{[-0.004, 0.063]}
\newcommand{\groundingGPTp}{0.083}
\newcommand{\groundingLlama}{+0.094}
\newcommand{\groundingLlamaCI}{[0.042, 0.150]}
\newcommand{\groundingClaude}{+0.002}
\newcommand{\groundingClaudeP}{0.756}

\newcommand{\mdeLlamaGrounding}{0.078}
\newcommand{\mdeGPTgrounding}{0.047}
\newcommand{\mdeClaudeGrounding}{0.017}

\newcommand{\mdeGPTrag}{0.099}
\newcommand{\mdeLlamaRag}{0.108}

\newcommand{\withinTaskSDlo}{0.065}
\newcommand{\withinTaskSDhi}{0.135}

\newcommand{\judgedGPT}{1410}
\newcommand{\judgedLlama}{1405}
\newcommand{\judgedClaude}{281}

\newcommand{\nNeededGPT}{130}

\newcommand{\nTasks}{47}
\newcommand{\nUsers}{9}
\newcommand{\nSims}{28}
\newcommand{\diffLowN}{4}
\newcommand{\diffMediumN}{25}
\newcommand{\diffHighN}{15}
\newcommand{\diffHardN}{3}
\newcommand{\domCalendar}{17}
\newcommand{\domCommunication}{9}
\newcommand{\domTravel}{10}
\newcommand{\domResearch}{11}

\newcommand{\nConflictAdhoc}{15}
\newcommand{\ragPerSim}{30}
\newcommand{\ragTotal}{840}
\newcommand{\ragPerUserRange}{60--150}

\newcommand{\judgeLowWeightFrac}{31\%}

\newcommand{\simVsRagAgree}{0.006}
\newcommand{\simVsRagGap}{0.12}

\newcommand{\dLlamaNoProfile}{-0.357} \newcommand{\mLlamaNoProfile}{0.112}
\newcommand{\dLlamaFlat}{-0.105}      \newcommand{\mLlamaFlat}{0.080}
\newcommand{\dLlamaRAG}{-0.210}       \newcommand{\mLlamaRAG}{0.108}
\newcommand{\dLlamaSimA}{-0.209}      \newcommand{\mLlamaSimA}{0.105}
\newcommand{\dLlamaSimG}{+0.094}      \newcommand{\mLlamaSimG}{0.078}
\newcommand{\dGPTNoProfile}{-0.367}   \newcommand{\mGPTNoProfile}{0.114}
\newcommand{\dGPTFlat}{-0.064}        \newcommand{\mGPTFlat}{0.069}
\newcommand{\dGPTRAG}{-0.205}         \newcommand{\mGPTRAG}{0.099}
\newcommand{\dGPTSimA}{-0.106}        \newcommand{\mGPTSimA}{0.055}
\newcommand{\dGPTSimG}{+0.028}        \newcommand{\mGPTSimG}{0.047}
\newcommand{\dClaudeNoProfile}{-0.418}\newcommand{\mClaudeNoProfile}{0.154}
\newcommand{\dClaudeFlat}{-0.030}     \newcommand{\mClaudeFlat}{0.058}
\newcommand{\dClaudeRAG}{-0.144}      \newcommand{\mClaudeRAG}{0.097}
\newcommand{\dClaudeSimA}{-0.066}     \newcommand{\mClaudeSimA}{0.066}
\newcommand{\dClaudeSimG}{+0.002}     \newcommand{\mClaudeSimG}{0.017}
\newcommand{\resolvedLlama}{5}
\newcommand{\resolvedGPT}{3}
\newcommand{\resolvedClaude}{2}

\newcommand{\headroomR}{+0.24}
\newcommand{\headroomT}{2.94}
\newcommand{\headroomN}{141}
\newcommand{\headroomLoHalf}{+0.022}
\newcommand{\headroomHiHalf}{+0.061}
\newcommand{\headroomNaiveR}{+0.48}

\newcommand{\loraBase}{Mistral-7B-Instruct-v0.3 (4-bit)}
\newcommand{\loraRank}{8}
\newcommand{\loraAlpha}{16}
\newcommand{\loraLR}{2\times10^{-5}}
\newcommand{\loraIters}{1{,}000}
\newcommand{\loraBatch}{4}
\newcommand{\loraLayers}{8}
\newcommand{\loraSeqLen}{2{,}048}
\newcommand{\loraNumAdapters}{3}

\newcommand{\loraN}{100}
\newcommand{\loraBaseR}{0.092}
\newcommand{\loraAmazonR}{0.098}
\newcommand{\loraSyntheticR}{0.109}
\newcommand{\loraPerUserR}{0.279}
\newcommand{\loraRoutedR}{0.306}
\newcommand{\loraBaseB}{0.792}
\newcommand{\loraAmazonB}{0.791}
\newcommand{\loraSyntheticB}{0.889}
\newcommand{\loraPerUserB}{0.911}
\newcommand{\loraRoutedB}{0.913}
\newcommand{\loraRoutingAcc}{72\%}
\newcommand{\loraAmazonVsBase}{+0.006}
\newcommand{\loraSyntheticVsBase}{+0.017}
\newcommand{\loraPerUserVsBase}{+0.187}
\newcommand{\loraRoutedVsBase}{+0.214}
\newcommand{\loraRoutedVsPerUser}{+0.027}
\newcommand{\loraRoutedVsPerUserCI}{[0.012, 0.042]}
\newcommand{\loraRoutedVsPerUserP}{0.0004}
\newcommand{\loraMDE}{0.022}

\newcommand{\octFlat}{0.765}
\newcommand{\octFlatArb}{0.788}
\newcommand{\octSim}{0.793}
\newcommand{\octSimVsFlatArb}{+0.005}
\newcommand{\octSimVsFlatArbCI}{[-0.029, 0.041]}
\newcommand{\octSimVsFlatArbP}{0.81}
\newcommand{\octFlatArbVsFlat}{+0.023}
\newcommand{\octFlatArbVsFlatCI}{[-0.007, 0.057]}
\newcommand{\octFlatArbVsFlatP}{0.15}
\newcommand{\octSimVsFlat}{+0.028}
\newcommand{\octSimVsFlatCI}{[-0.005, 0.064]}
\newcommand{\octSimVsFlatP}{0.10}
\newcommand{\simDrift}{0.035}
\newcommand{\simDriftP}{0.005}
\newcommand{\claudeFlatArb}{0.962}
\newcommand{\claudeFlatArbVsSim}{+0.008}
\newcommand{\gridPlans}{560}
\newcommand{\gridPlansTotal}{563}
\newcommand{\gridSpearman}{0.727}
\newcommand{\gridKappa}{0.345}
\newcommand{\gridMAE}{0.243}
\newcommand{\gridOffset}{0.184}

\newcommand{\dnTasks}{20}
\newcommand{\dnSimDrop}{0.056}
\newcommand{\dnSimGDrop}{0.116}
\newcommand{\dnNoProfDrop}{0.015}
\newcommand{\dnGroundWith}{+0.111}
\newcommand{\dnGroundWithout}{+0.051}
\newcommand{\nGoldTasks}{12}
\newcommand{\nParseFail}{6}
\newcommand{\nGenTotal}{3{,}102}
\newcommand{\mdeMin}{0.017}
\newcommand{\mdeMax}{0.154}

\newcommand{\octGapAbs}{0.005}
\newcommand{\arbitrationLlamaAbs}{0.209}

\newcommand{\simguide}{SimGuide}
\newcommand{\simbench}{SimBench}
\newcommand{\simspace}{\mathcal{K}}
\newcommand{\simset}{\mathcal{S}}

\title{\simguide{}: Typed Multi-Context User Representations\\ for Preference-Conditioned Agent Planning}

\author{Chirag Shah}
\affiliations{University of Washington, Seattle, USA\\
\texttt{chirags@uw.edu}}

\begin{document}
\maketitle

\begin{figure*}[t]
\centering
\includegraphics[width=\linewidth]{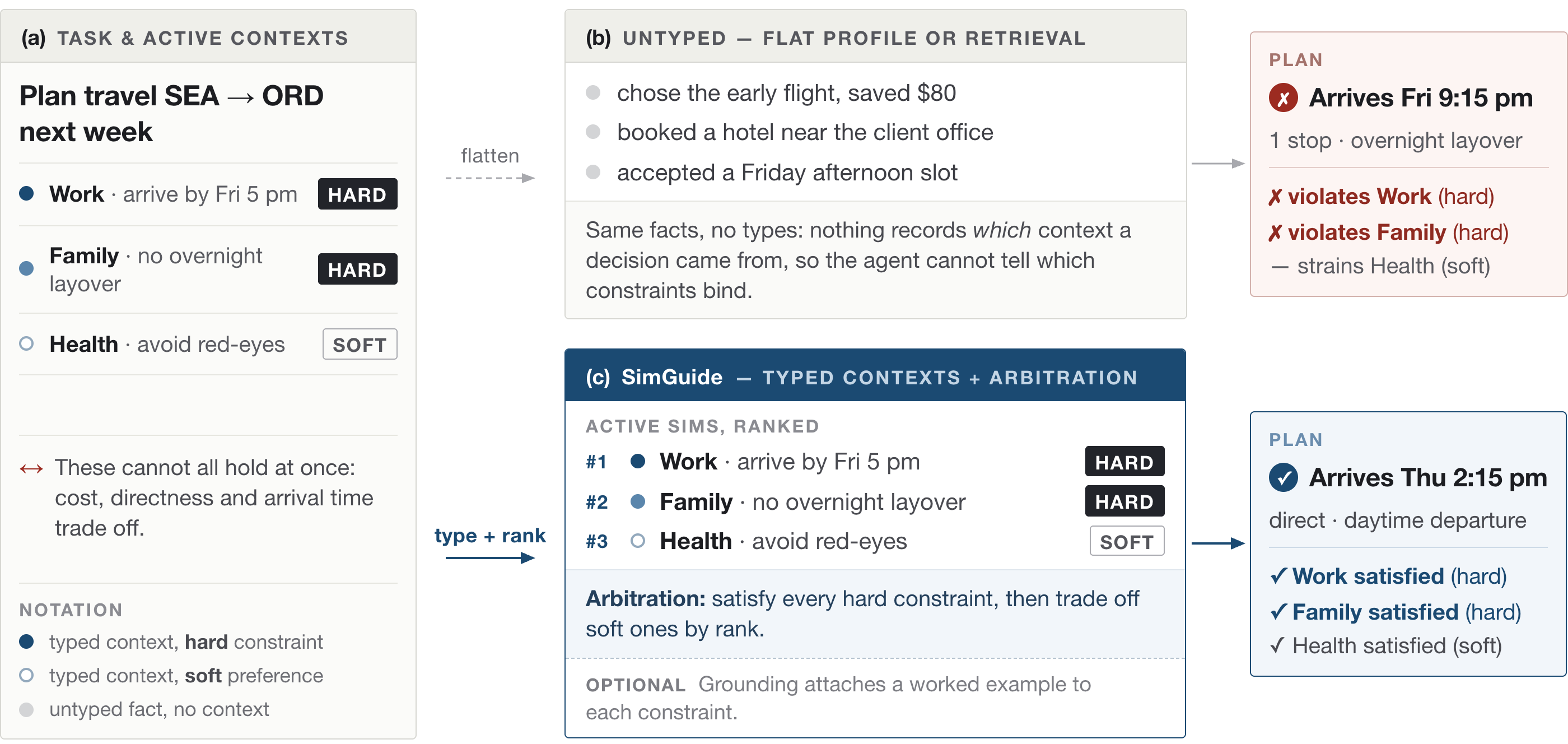}
\caption{\textbf{Preference-conditioned planning under conflicting user contexts.}
(a)~One task activates three contexts whose constraints pull in different directions.
(b)~Flat profiles and retrieval over past decisions supply the same facts without recording which
context each belongs to, so the agent cannot tell which constraints bind; the plan shown breaks the
Work \textsc{hard} deadline.
(c)~Typing the contexts and ranking them makes the binding constraints explicit and the trade-off
resolvable. Procedural grounding attaches a worked example to each constraint and is an optional
component whose benefit depends on model headroom.}
\label{fig:overview}
\end{figure*}

\begin{abstract}
Agents that act on a user's behalf must plan differently for different users, and increasingly do
so from some structured representation of user context and not from raw interaction history.
How much that structure is worth, and which parts of it carry the value, is largely unmeasured. We
introduce \simbench{}, \nTasks{} preference-conditioned planning tasks over \nUsers{} synthetic
users represented as \nSims{} typed, potentially conflicting context blocks, where the correct plan
depends on which contexts are active and how their conflicts are resolved. Against it we evaluate
\simguide{}, a framework combining typed multi-context representation, explicit conflict
arbitration, and optional procedural grounding of individual constraints. Across three models and
six user-context representations, \simguide{}'s typed blocks with arbitration outperform retrieval
over the same user's past decisions by \simVsRagLlama{},
\simVsRagGPT{} and \simVsRagClaude{} Preference Adherence on Llama~3.3~70B, GPT-4o and Claude
Sonnet~4.5 respectively (all $p<0.001$), and removing the arbitration instruction from typed
blocks costs up to \arbitrationLlamaAbs{}. The typing itself adds nothing we can resolve once arbitration is held
fixed: a flat profile given the same arbitration instruction scores \octFlatArb{} against
\octSim{} for typed Sims on GPT-4o in a same-day run ($p=\octSimVsFlatArbP{}$). Grounding each constraint with a worked example of past application
helps only where the model has headroom: \groundingLlama{} on Llama~70B ($p<0.001$), falling to
\groundingGPT{} on GPT-4o and \groundingClaude{} on Claude, which already scores perfectly on
\ceilClaudeSim{} of \nTasks{} tasks without it. We report the benchmark's minimum detectable effect
alongside its results. The benchmark ships with a provenance audit that re-derives every reported
number from the prompt that produced it.
\end{abstract}

\section{Introduction}
\label{sec:intro}

A single user is not a single set of preferences. The same person books travel differently for a
client engagement than for a family trip, and an assistant that flattens those contexts into one
profile will confidently satisfy the wrong one. The assistant
may hold every relevant fact about the user and still be unable to determine which facts apply,
because a flat profile records what a user prefers without recording \emph{when}. Retrieval over
past behaviour has the same gap in a different form. It surfaces decisions that are similar to the
current situation, but carries no representation of which life context those decisions belonged to,
and therefore cannot say what happens when two contexts pull in opposite directions.

\simguide{} takes the alternative position (Figure~\ref{fig:overview}): represent a user as a set of typed, domain-specific
context blocks (\emph{Sims}, following \citet{shah2025agents}), each carrying its own priorities,
hard and soft constraints, and communication style, together with an explicit rule for arbitrating
between them when they conflict. Optionally, each constraint may be grounded with a worked example
of the user applying it in a past decision. This yields a design space,
and the question this paper asks is empirical: which parts of it actually matter, and by how much?

Answering that requires tasks where the correct plan depends on which contexts are
active. Existing agent benchmarks evaluate task completion against fixed policies, where a
context-blind agent can succeed without knowing anything about the user; existing personalization
benchmarks test preference inference within a single context per instance, so nothing conflicts and
arbitration is never exercised. We therefore construct \simbench{}, in which each task activates a
subset of a user's Sims and admits different correct plans depending on that subset.

Our results give a clear ordering. Whether the agent receives structured user context at all
dominates everything else. Among representations, typed blocks with arbitration beat retrieval over
past decisions on all three models tested, with the retrieval corpus generated from the same
underlying preferences. Within that advantage, the arbitration instruction matters most. Removing
it from typed blocks costs up to \arbitrationLlamaAbs{} PA, while a flat profile given the same instruction scores
within \octGapAbs{} of typed Sims in a same-day run. Once arbitration is held fixed, typing
adds nothing the benchmark can resolve. Procedural grounding, by contrast, pays only where the
model has room to improve, and vanishes on a frontier model that already solves most of the
benchmark from typed constraints alone.

That last finding is a correction of our own earlier work, and we treat it as a methodological
result as much as an empirical one. An earlier version of this study reported grounding as an
unconditional benefit. It did not survive a corrected evaluation pipeline. The defect class was
mundane and, we suspect, common: silent fallbacks that turned missing data into plausible-looking
prompts, and no mechanism connecting a number in a table to the string that produced it. We
therefore ship a provenance audit that re-derives every reported result from its prompt, and we report the benchmark's minimum detectable effect before its results, so that
effects too small to resolve are labelled unresolved, not absent.

\paragraph{Contributions.}
\begin{enumerate}
\item \simbench{}: \nTasks{} preference-conditioned planning tasks over \nUsers{} users and
      \nSims{} typed Sims, where the correct plan depends on which contexts are active, with a
      provenance audit released alongside it.\footnote{Benchmark, evaluation code and
      provenance audit are distributed with this preprint as ancillary files.}
\item A decomposition of the \simguide{} design space showing that explicit arbitration and
      procedural grounding carry the gain, while typing adds nothing resolvable once arbitration is
      held fixed; typed Sims with arbitration beat retrieval over the same user's past decisions on
      all three models tested.
\item Evidence that procedural grounding is conditional on model headroom, with a saturation
      analysis showing where the benchmark can and cannot resolve representation effects.
\end{enumerate}
\noindent We also report an exploratory adapter study (Exploratory: Parametric Personalization), which we do not
count as a contribution because it is scored on surface similarity, not on Preference Adherence.

\section{Related Work}

\paragraph{Memory and personalization in language models.}
Memory-augmented LLM systems span a wide design space. RAG~\citep{lewis2020rag} grounds responses in retrieved documents but treats retrieval and generation as independent steps, with no representation of what user contexts the retrieved decisions belong to. MemGPT~\citep{packer2023memgpt} introduces tiered memory management for extended conversations. LaMP~\citep{salemi2024lamp} established a benchmark for personalized NLP tasks using historical user data; PersonalLLM~\citep{zollo2025personalllm} frames personalization as a meta-learning problem over diverse preference distributions. LOCOMO~\citep{maharana2024locomo} evaluates very long-term conversational memory and reports that long-context models and retrieval augmentation both improve over base LLMs while still lagging well behind human performance. Recent work on preference inference from agent trajectories~\citep{aroca2024predict} similarly finds that generic preference representations fail to capture individual constraints, motivating structured decomposition. Closer to our setting, PrefEval~\citep{zhao2025prefeval} tests whether LLMs follow stated user preferences across long multi-session conversations, ETAPP~\citep{hao2025etapp} evaluates personalized tool invocation against user profiles, AppAgent-Pro~\citep{zhao2025appagentpro} builds a proactive GUI agent that integrates information across app domains, and SmartAgent~\citep{zhang2026smartagent} reasons about user preferences while acting in GUI environments. These systems personalize agent behaviour, but each task applies one user's preferences without arbitrating between several contexts that are active at once and conflict. Our novelty claim is limited to that case: explicit arbitration across simultaneously active user contexts. \simguide{} represents users as typed, multi-context Sims carrying an explicit arbitration rule, and we measure how much of the benefit each component supplies.

\paragraph{Parametric personalization.}
OPPU~\citep{tan2024oppu} assigns each user a dedicated LoRA adapter~\citep{hu2021lora}, outperforming prompt-based methods on LaMP. Per-Pcs~\citep{tan2024perpcs} addresses OPPU's scalability by decomposing user adapters into shareable components assembled via gating. Profile-to-PEFT~\citep{tan2025p2p} uses a hypernetwork to map an encoded user profile directly to adapter parameters, removing per-user training at deployment. LLM-Personalize~\citep{han2024llmpersonalize} extends parametric personalization to robot task planners via self-training, but operates within a single user context without conflict arbitration across multiple life domains. These methods condition adaptation on unstructured user history embeddings and assume that behavioral training data is available in the evaluation domain. \simguide{} instead conditions adaptation on typed, multi-context representations, which permits interpretable conflict resolution and context-dependent adapter selection; our parametric results point to Sim type, rather than user identity, as the better granularity at realistic per-user data volumes.

\paragraph{Agent benchmarks.}
WebArena~\citep{zhou2024webarena} provides realistic web environments with 812 tasks. AgentBench~\citep{liu2024agentbench} evaluates 29 LLMs across 8 environments. $\tau$-bench~\citep{yao2024tau} is a simulation framework for evaluating customer service agents in real-world domains (airline and retail), where an agent must navigate dynamic user interactions and execute API calls subject to domain-specific policy guidelines. $\tau^2$-bench~\citep{barres2025tau2} extends this to dual-control settings. OSWorld~\citep{xie2024osworld} extends evaluation to full operating systems. The shared limitation: none of these test whether agents adapt their planning based on persistent, multi-context user preferences. \simbench{}, introduced below, is designed around this property.

\paragraph{Sims architecture.}
\citet{shah2025agents} propose an ecosystem of agents, assistants, and Sims, where Sims are structured representations of user preferences across life contexts. The concept shares motivation with generative agent simulations~\citep{park2023generative}, which also model persistent, context-dependent behavior, but Sims are designed for personalized planning rather than social simulation. The present paper provides the missing empirical evaluation: it separates the contributions of typing, arbitration and procedural grounding across three models on a purpose-built benchmark, quantifies what that benchmark can and cannot resolve, and reports an exploratory study of parametric personalization via LoRA fine-tuning.

\section{The \simguide{} Framework}
\label{sec:method}

\simguide{} is a framework for supplying user context to a planning agent, with three components:
a typed multi-context representation, an explicit arbitration rule over conflicting contexts, and
optional procedural grounding of individual constraints. Our experiments treat these as separable
design choices and measure each.

\subsection{Sim Representation}

Following \citet{shah2025agents}, a \emph{Sim} is a typed, structured object representing a user in
one life context. For user $u$ and context $k \in \simspace_u \subseteq \simspace$:

\begin{equation}
    S_k = \langle P_k,\ C_k,\ \sigma_k,\ E_k \rangle
    \label{eq:sim}
\end{equation}

\noindent In Equation~\ref{eq:sim}, $P_k = \langle (\pi_i, w_i) \rangle_{i=1}^{|P_k|}$ is an ordered sequence of
priorities $\pi_i$ with weights $w_i \in [0,1]$; $C_k = \{(\gamma_j, \mathrm{sev}_j)\}$ is a set of
constraints $\gamma_j$ each labelled $\mathrm{sev}_j \in \{\textsc{hard}, \textsc{soft}\}$;
$\sigma_k$ is a communication-style descriptor; and $E_k$ is a set of procedural examples
(defined below). The user's Sim set is $\simset_u = \{S_k : k \in \simspace_u\}$,
where $\simspace_u$ is the subset of contexts defined for that user, not the full space.

Hard constraints must never be violated. Soft constraints may be traded off under explicit
justification. This typing defines correctness: a plan violating a hard constraint is
incorrect regardless of other merits.

\subsection{Arbitration}
\label{sec:arbitration}

Given a task $t$ with active Sims $A_t \subseteq \simset_u$ ordered by priority, and abstract tools
$\mathcal{T}$, the agent produces a plan $\rho = (s_1, \ldots, s_n)$ with $s_i = (\tau_i,
\theta_i)$, where $\tau_i \in \mathcal{T}$ is a tool and $\theta_i$ its parameters.

We formalise the objective that the arbitration instruction expresses. Let $\Pi$ be the set of
plans and order the active Sims by priority rank, $S_{(1)}, \ldots, S_{(m)}$, with rank 1 highest.
Hard constraints first restrict the plan set:

\begin{equation}
\begin{split}
    \Pi_0 = \{\rho \in \Pi :\ & \mathrm{sat}(\rho, \gamma_j)\ \text{for every} \\
    & (\gamma_j, \textsc{hard}) \in C_k,\ S_k \in A_t\}
\end{split}
    \label{eq:feasibility}
\end{equation}

\noindent We assume the active hard constraints are jointly feasible, so $\Pi_0$ is non-empty. Each Sim then scores a plan by its satisfied
priorities less a penalty for violated soft constraints,
$U_k(\rho) = \sum_{(\pi_i, w_i) \in P_k} w_i\, \mathbb{1}[\mathrm{sat}(\rho, \pi_i)]
- \lambda \sum_{(\gamma_j, \textsc{soft}) \in C_k} (1 - \mathbb{1}[\mathrm{sat}(\rho, \gamma_j)])$,
and the Sims are applied in rank order:

\begin{equation}
    \Pi_r = \operatorname*{arg\,max}_{\rho \in \Pi_{r-1}} U_{(r)}(\rho), \quad r = 1, \ldots, m,
    \qquad \rho^* \in \Pi_m
    \label{eq:arbitration}
\end{equation}

\noindent Equation~\ref{eq:feasibility} makes the hard constraints of every active Sim inviolable.
Equation~\ref{eq:arbitration} is lexicographic: a lower-ranked Sim only breaks ties among plans that
are already best for every Sim above it, so no amount of lower-ranked preference can outweigh a
higher-ranked one. The agent is not given this objective as an algorithm. It is told the Sims'
priority order and asked to resolve conflicts by it, and the equations state what a correct
resolution looks like.

\subsection{Procedural Grounding}
\label{sec:grounding}

A declarative constraint states a rule; it does not state how the user applies it. We test whether
supplying that application helps. For each constraint $\gamma_j$ we optionally attach a worked
example $e_j$ describing a past decision in which the user applied it:

\begin{quote}
\textbf{Declarative:} ``No flights departing after 8:00\,PM (\textsc{hard}).''

\textbf{Procedurally grounded:} ``No flights departing after 8:00\,PM (\textsc{hard}).
\emph{When booking SEA$\to$ORD last month, rejected the 9:30\,PM departure (\$120 cheaper) and
booked the 2:15\,PM flight, accepting a longer layover to protect sleep before a morning client
meeting.}''
\end{quote}

A second case shows the complementary situation, where the constraint is satisfiable in many ways
and the example reveals which the user prefers:

\begin{quote}
\textbf{Declarative:} ``Arrive before the 9:00\,AM standup (\textsc{soft}).''

\textbf{Procedurally grounded:} ``Arrive before the 9:00\,AM standup (\textsc{soft}).
\emph{Chose a 6:40\,AM arrival over a 8:35\,AM one on the last two trips, treating a 90-minute
buffer as the working minimum after a delayed connection caused a missed standup in March.}''
\end{quote}

Here the declarative form underdetermines the plan: any arrival before 9:00\,AM satisfies it, while
the example identifies the buffer the user actually applies. The grounded representation is
$S_k^{+} = \langle P_k, C_k, \sigma_k, \{(\gamma_j, e_j) : \gamma_j \in C_k\} \rangle$.

Two mechanisms could make this help. Examples may be easier to follow than stated rules, consistent
with findings that models track demonstrations more reliably than
instructions~\citep{brown2020gpt3,min2022rethinking}. And examples implicitly encode the user's
trade-off reasoning, which may transfer to situations the constraint does not literally cover. We
report below that the benefit is real but conditional on the model having
headroom, and we do not claim it is unconditional.

All models receive the same grounded template with the same exemplar density; we deliberately avoid
model-specific grounding variants, since a condition that differs by model cannot support a
cross-model comparison.

\subsection{Experimental Conditions}

We evaluate six conditions varying only the user context supplied to the agent
(Table~\ref{tab:conditions}), plus a seventh, Flat+Arb, that gives the flat profile the same
two-line arbitration instruction as Sim. Flat+Arb separates typing from arbitration, which the
Flat--Sim contrast confounds. It was added after the main grid and is run in a separate same-day
comparison (see Results). The task, tool schema, and output contract are identical throughout.

\begin{table}[t]
\caption{Experimental conditions. All receive the same task, tools and output contract, and
differ only in the form of the user context.}
\label{tab:conditions}
\centering
\footnotesize
\setlength{\tabcolsep}{4pt}
\begin{tabular}{@{}lcccc@{}}
\toprule
Condition & Context & Typed & Arbitr. & Ground. \\
\midrule
No-Profile & none      & --- & --- & --- \\
Flat       & merged    & no  & no  & no  \\
Flat+Arb   & merged    & no  & yes & no  \\
RAG        & retrieved & no  & no  & --- \\
Sim-A      & Sims      & yes & no  & no  \\
Sim        & Sims      & yes & yes & no  \\
Sim+G      & Sims      & yes & yes & yes \\
\bottomrule
\end{tabular}
\end{table}

The RAG condition is the substantive baseline. For each user we generate a corpus of \ragPerSim{}
synthetic past decisions per Sim (\ragPerUserRange{} per user, \ragTotal{} total), written from that
Sim's priorities and constraints but without reference to any benchmark task, so the corpus cannot
leak gold plans. At inference we embed the corpus and the task description with all-MiniLM-L6-v2 and
inject the top five decisions by cosine similarity~\citep{reimers2019sbert}. The Sim each decision came from is deliberately
withheld: retrieval over untyped behaviour is the baseline that typing is compared against, so
surfacing the type would leak the treatment into the control.

\section{\simbench{}}
\label{sec:simbench}

\simbench{} is built around one property: the correct plan depends on which user context is active,
not only on the task description. Existing benchmarks do not test this. Agent benchmarks such as
WebArena and $\tau$-bench evaluate task completion against fixed policies, where a context-blind
agent can succeed without knowing anything about the user. Personalization benchmarks such as LaMP
and LOCOMO test preference inference within a single context per instance, so nothing conflicts and
arbitration is never exercised. \simbench{} tasks are constructed so that a plan satisfying one
active Sim can simultaneously violate another, and so that a No-Profile agent has no signal in the
task description from which to recover.

\paragraph{Composition.}
The release contains \nTasks{} tasks over \nUsers{} canonical synthetic users comprising \nSims{}
Sims, each user holding 2--5 contexts. Domains are calendar (\domCalendar{}), research
(\domResearch{}), travel (\domTravel{}) and communication (\domCommunication{}). Difficulty is
annotated low (\diffLowN{}), medium (\diffMediumN{}), high (\diffHighN{}) and hard (\diffHardN{}).
Correct tool parameters require user constraints that are absent from the task description.

\paragraph{Design notes.}
\dnTasks{} of the \nTasks{} task descriptions end with an authoring note describing how a
context-blind agent would fail, and several of these notes name the user constraint at stake. Every
condition saw them, No-Profile included, so they cannot favour one condition by construction, but
they can change effect sizes. We checked by rerunning the \dnTasks{} tasks with and without the
notes on GPT-4o ($k=1$, same day). Removing them lowers Sim by \dnSimDrop{}, Sim+G by
\dnSimGDrop{}, and No-Profile by \dnNoProfDrop{}. The grounding gain on these tasks falls from
\dnGroundWith{} ($p_{\text{Holm}}<0.001$) to \dnGroundWithout{} (not significant). The reported
grid keeps the notes; SimBench v1.2 removes them (see Follow-up work under Limitations).

\paragraph{Conflict annotation.}
Each task is annotated with the conflict categories it contains. Four canonical categories recur
across the suite: priority, constraint, temporal and communication conflicts. A further
\nConflictAdhoc{} task-specific labels appear, mostly on single tasks, describing narrower
conflicts such as venue or provider selection. We report the annotation as released, without
collapsing it, and stratified analyses use the canonical four.

\paragraph{Metrics.}
We report one metric. \emph{Preference Adherence} (PA) measures constraint satisfaction and is
scored by an LLM judge given the task, the active Sim definitions, and the plan. The release also
defines two deterministic metrics, Plan Correctness (tool selection and parameter keys against a
gold plan) and Conflict Resolution Accuracy (the declared conflict strategy against a gold
strategy), but we do not report them. Only \nGoldTasks{} of \nTasks{} tasks carry a gold plan and a
gold strategy, and CRA as implemented compares the plan's strategy label with free-text gold
strategies, so it scores 0 for every plan. The fix is to map the gold strategies to the plan
schema's labels and to add a deterministic hard-constraint checker, and we will report both metrics
once they are defined on every task.

\paragraph{Judge reliability.}
The judge is GPT-4o. Because model-based judges carry known biases~\citep{zheng2023judging} we
report measured reliability rather than assuming it. A second judge, Claude Sonnet~4.6, scored the
first sample of every task and condition for the GPT-4o and Llama planners in the reported grid
(\gridPlans{} of \gridPlansTotal{} plans; three verdicts were malformed). Per-plan Spearman
correlation is \gridSpearman{}, Cohen's $\kappa$~\citep{cohen1960kappa} on per-constraint verdicts
is \gridKappa{}, and mean absolute difference in PA is \gridMAE{}, with GPT-4o scoring higher by
\gridOffset{} on average. The two judges rank the six conditions identically for both planners
(Kendall $\tau=1.0$). Separately,
the judge weights each constraint by an estimate of its relevance to the task, and
\judgeLowWeightFrac{} of constraints receive low weight from a lexical-overlap heuristic. Both facts
bound the resolution of any PA-based comparison, and we treat differences smaller than the
per-comparison detection threshold as unresolved. Human-annotated agreement remains future work.

\section{Experimental Design}
\label{sec:design}

\paragraph{Models.}
We evaluate three models spanning a wide capability range: GPT-4o, Claude Sonnet~4.5, and
Llama~3.3~70B Instruct~\citep{openai_gpt4o_system_card_2024,anthropic_claude_4_5_sonnet_model,meta2024llama3}.
A fourth arm using a small open-source model (Llama~8B class) was planned
and is absent for an infrastructure reason we record for reproducibility: during the study the
provider moved all 8B-class models off its serverless tier, requiring a provisioned dedicated
endpoint, while continuing to serve Llama~3.3~70B on the same key. The three remaining models span
baseline Preference Adherence from \llamaSim{} to \claudeSim{} under typed Sims, which is the range
the analysis below relies on.

\paragraph{Protocol.}
GPT-4o and Llama~3.3~70B are run with $k=\kSamples{}$ independent samples per task at temperature
\samplingTemp{} and fixed seeds. Claude is run at $k=1$ and temperature 0: it is included as a saturation probe, and
additional samples buy nothing on a model that already reaches ceiling on
\ceilClaudeSim{}/\nTasks{} tasks. The unit of analysis is the per-task mean over samples, not the
individual generation. Comparisons are paired at the task level and tested two ways, a paired
bootstrap over tasks (10{,}000 resamples)~\citep{efron1993bootstrap} and a Wilcoxon signed-rank
test~\citep{wilcoxon1945}, with Holm correction~\citep{holm1979} within each comparison family. All models render an identical Sim+G template with a mean of 4.17
procedural exemplars per prompt; an earlier provider-conditional branch that supplied compact
grounding to one model only has been removed, since a condition must mean the same thing across
models for a cross-model comparison to be interpretable. Mean within-task standard deviation ranges from \withinTaskSDlo{}
to \withinTaskSDhi{} PA across the sampled cells.

\paragraph{Provenance.}
Every reported result is verified against the prompt that produced it. Generations are cached under
a hash of provider, model, rendered prompt, plan schema, and sampling parameters; an audit
re-renders each prompt, recomputes the hash, and confirms the stored response is byte-identical to
the cached generation. A run that does not reproduce is treated as unattributable and regenerated.
The audit additionally rejects any prompt containing a placeholder or missing-data marker, any task
whose Sim identifiers fail to resolve, and any grounded prompt lacking procedural examples. Of
\nGenTotal{} generations, \nParseFail{} failed to parse (five from Llama, one from Claude) and are
excluded; all \judgedGPT{}, \judgedLlama{} and \judgedClaude{} parsed plans were scored. The audit ships with the benchmark; its checks are not specific to our
system.

\paragraph{What this design can detect.}
Detection thresholds depend on the variance of the specific contrast, so we compute a minimum
detectable effect (MDE) at 80\% power for each comparison~\citep{card2020power,dror2018hitchhiker}
instead of pooling across them; pooling
inflates the threshold because the No-Profile contrast carries far more variance than the others. At
$n=\nTasks{}$ tasks the resulting MDEs range from \mdeMin{} to \mdeMax{} PA depending
on model and contrast. We report each alongside its estimate, and describe effects below their own
threshold as unresolved rather than absent. Two consequences follow: the
retrieval comparisons are comfortably resolved on every model (MDE \mdeGPTrag{} on GPT-4o,
\mdeLlamaRag{} on Llama~70B), while the grounding comparison is resolved only on Llama~70B
(\mdeLlamaGrounding{}). Note also that saturation shrinks variance and therefore shrinks the MDE:
Claude's thresholds are the smallest in the grid because most of its comparisons are close
to degenerate, so for that model the ceiling counts in Table~\ref{tab:main} are more informative
than the thresholds.

\section{Results}
\label{sec:results}

\begin{table}[t]
\caption{Preference Adherence on \simbench{}, $n=\nTasks{}$ tasks. Llama and GPT-4o use
$k=\kSamples{}$ samples per task; Claude uses $k=1$ as a saturation probe. Models are ordered by
baseline capability. Bold marks a best condition only where its margin over Sim exceeds that
contrast's detection threshold (Table~\ref{tab:mde}). The final row counts tasks scoring
PA${}=1.000$ under typed Sims.}
\label{tab:main}
\centering
\small
\setlength{\tabcolsep}{4pt}
\begin{tabular}{lccc}
\toprule
Condition & Llama 3.3 70B & GPT-4o & Claude 4.5 \\
\midrule
No-Profile             & \llamaNoProfile{} & \gptNoProfile{} & \claudeNoProfile{} \\
RAG                    & \llamaRAG{}       & \gptRAG{}       & \claudeRAG{} \\
Sim-A (no arbitration) & \llamaSimA{}      & \gptSimA{}      & \claudeSimA{} \\
Flat                   & \llamaFlat{}      & \gptFlat{}      & \claudeFlat{} \\
Sim                    & \llamaSim{}       & \gptSim{}       & \claudeSim{} \\
Sim+G                  & \textbf{\llamaSimG{}} & \gptSimG{}      & \claudeSimG{} \\
\midrule
Tasks at ceiling (Sim) & \ceilLlamaSim{}/\nTasks{} & \ceilGPTsim{}/\nTasks{} & \ceilClaudeSim{}/\nTasks{} \\
\bottomrule
\end{tabular}
\end{table}

\subsection{Structured Representations Beat Retrieval}
The central comparison is between typed, arbitrated user context and retrieval over the same
user's past decisions. Structure wins on every model: \simVsRagLlama{} PA on Llama~3.3~70B
(95\% CI \simVsRagLlamaCI{}), \simVsRagGPT{} on GPT-4o (CI \simVsRagGPTci{}), and
\simVsRagClaude{} on Claude Sonnet~4.5 (CI \simVsRagClaudeCI{}), all $p<0.001$ after Holm
correction. Every estimate exceeds the design's detection threshold, and the two sampled models
agree to within \simVsRagAgree{} PA despite a \simVsRagGap{} gap in absolute performance.

This is a like-for-like comparison. The retrieval baseline draws from a corpus of \ragPerSim{}
synthetic past decisions per Sim (\ragPerUserRange{} per user, \ragTotal{} total), embedded with
all-MiniLM-L6-v2 and injected as the top five by cosine similarity to the task description. Those
decisions are generated from the same Sim definitions the structured conditions receive, so the
conditions differ in the \emph{form} of the user information, not in its source or quantity.

\subsection{Arbitration Separates Sims from Flat Profiles}
In the main grid, flat concatenation of identical Sim content underperforms typed blocks by
\simVsFlatLlama{} PA on Llama~70B ($p<0.001$), \simVsFlatGPT{} on GPT-4o ($p=\simVsFlatGPTp{}$) and
\simVsFlatClaude{} on Claude ($p=\simVsFlatClaudeP{}$). The GPT-4o and Claude estimates fall below
their detection thresholds (\mGPTFlat{} and \mClaudeFlat{}). This contrast does not isolate typing,
however. Flat differs from Sim in two ways: it drops the typed blocks, and it drops the arbitration
instruction.

Flat+Arb removes the second difference. We ran it on GPT-4o ($k=\kSamples{}$, temperature
\samplingTemp{}, seed \samplingSeed{}) on the same day as fresh runs of Flat and Sim, so that all
three cells share one generator state; we do not compare it with the main-grid cells. Flat scores
\octFlat{}, Flat+Arb \octFlatArb{}, and Sim \octSim{}. Sim leads Flat+Arb by \octSimVsFlatArb{}
(95\% CI \octSimVsFlatArbCI{}, $p=\octSimVsFlatArbP{}$), Flat+Arb leads Flat by \octFlatArbVsFlat{}
(CI \octFlatArbVsFlatCI{}, $p=\octFlatArbVsFlatP{}$), and Sim leads Flat by \octSimVsFlat{} (CI
\octSimVsFlatCI{}, $p=\octSimVsFlatP{}$). None of the three is resolved. On Claude ($k=1$,
temperature 0), Flat+Arb scores \claudeFlatArb{} against \claudeSim{} for Sim, a difference of
\claudeFlatArbVsSim{} in favour of the flat profile. Once arbitration is held fixed, typing adds
nothing the benchmark can resolve on either model. We could not run Flat+Arb on Llama~70B because
the provider stopped serving the model on its serverless tier, so the Llama Flat--Sim gap remains
a joint effect of typing and arbitration.

The fresh Sim run on GPT-4o also scores below its main-grid value (\octSim{} against \gptSim{}, a
drop of \simDrift{}, $p=\simDriftP{}$), while Flat is unchanged (\octFlat{} against \gptFlat{}). We
attribute the gap to drift in the served model between July and October, since prompts, sampling
parameters and seeds are identical. This is why we report the Flat+Arb comparison only within its
own same-day run.

Removing the arbitration instruction while retaining typed blocks costs \arbitrationLlama{} PA on
Llama~70B, \arbitrationGPT{} on GPT-4o, and \arbitrationClaude{} on Claude, all $p<0.001$. The
instruction to resolve conflicts by priority order matters most for the weakest model, as the representation effects do.

Every populated representation beats no profile at all by a wide margin (\profileVsNoneLlama{},
\profileVsNoneGPT{}, \profileVsNoneClaude{} for typed Sims). Whether the agent receives structured
user context remains the dominant variable; the choice among formats is second-order, though not
negligible.

Taken together, these contrasts place the gain in arbitration. Typed blocks
without the instruction lose to typed blocks with it on every model, and a flat profile with the
instruction does as well as typed blocks on the two models where we could test it. The framework's
third component, procedural grounding, behaves differently.

\subsection{Procedural Grounding Scales Inversely with Headroom}

Grounding each constraint with a worked example of the user applying it produces
\groundingLlama{} PA on Llama~3.3~70B (CI \groundingLlamaCI{}, $p=\groundingLlamaP{}$), \groundingGPT{} on
GPT-4o (CI \groundingGPTci{}, $p=\groundingGPTp{}$), and \groundingClaude{} on Claude
($p=\groundingClaudeP{}$). Only the first exceeds its detection threshold
(\mdeLlamaGrounding{}); the GPT-4o and Claude estimates fall below theirs (\mdeGPTgrounding{} and
\mdeClaudeGrounding{}) and we report them as unresolved. Part of the GPT-4o estimate also depends
on the design notes: on the \dnTasks{} tasks that carry them, the grounding gain falls from
\dnGroundWith{} to \dnGroundWithout{} when the notes are removed (see Design notes in the \simbench{} section). The ordering is monotone in the models'
baseline performance under typed Sims: \llamaSim{}, \gptSim{}, \claudeSim{}.

We read this as a matter of headroom. The final row
of Table~\ref{tab:main} makes the constraint explicit: Claude already scores PA${}=1.000$ on
\ceilClaudeSim{} of \nTasks{} tasks under plain typed Sims, leaving almost nothing for a
representation change to recover, while Llama~70B is at ceiling on only \ceilLlamaSim{}. Grounding
pays where the model has room to improve, and \simbench{} supplies that room for mid-capability
models but not for frontier ones.

\begin{table}[t]
\caption{Effect\,/\,detection threshold for every contrast against Sim, in PA points. Each cell
gives the observed difference and the minimum detectable effect for that specific comparison at
80\% power. \textit{Italic} marks effects smaller than their own threshold, which we mark
unresolved. The benchmark resolves every contrast on the weakest model and
fewer than half on the strongest. Note that MDE is a prospective property of the design while the
significance tests are retrospective, so the two can disagree at the margin: Claude's Sim-A
contrast is significant under the paired tests yet sits exactly at its threshold, and we treat it
as unresolved.}
\label{tab:mde}
\centering
\small
\setlength{\tabcolsep}{4pt}
\begin{tabular}{lccc}
\toprule
vs.\ Sim & Llama 3.3 70B & GPT-4o & Claude 4.5 \\
\midrule
No-Profile & \dLlamaNoProfile{}\,/\,\mLlamaNoProfile{} & \dGPTNoProfile{}\,/\,\mGPTNoProfile{} & \dClaudeNoProfile{}\,/\,\mClaudeNoProfile{} \\
RAG        & \dLlamaRAG{}\,/\,\mLlamaRAG{}            & \dGPTRAG{}\,/\,\mGPTRAG{}            & \dClaudeRAG{}\,/\,\mClaudeRAG{} \\
Sim-A      & \dLlamaSimA{}\,/\,\mLlamaSimA{}          & \dGPTSimA{}\,/\,\mGPTSimA{}          & \textit{\dClaudeSimA{}\,/\,\mClaudeSimA{}} \\
Flat       & \dLlamaFlat{}\,/\,\mLlamaFlat{}          & \textit{\dGPTFlat{}\,/\,\mGPTFlat{}} & \textit{\dClaudeFlat{}\,/\,\mClaudeFlat{}} \\
Sim+G      & \dLlamaSimG{}\,/\,\mLlamaSimG{}          & \textit{\dGPTSimG{}\,/\,\mGPTSimG{}} & \textit{\dClaudeSimG{}\,/\,\mClaudeSimG{}} \\
\midrule
resolved   & \resolvedLlama{}/5 & \resolvedGPT{}/5 & \resolvedClaude{}/5 \\
\bottomrule
\end{tabular}
\end{table}

If headroom is really what governs the effect, it should appear within a model and not only
across the three of them. It does. Pooling all \headroomN{} task-by-model cells and estimating each
task's headroom as $1 - \tfrac{1}{2}(\mathrm{PA}_{\text{Flat}} + \mathrm{PA}_{\text{Sim-A}})$, from two conditions that share no measurement noise with either
term of the Sim+G minus Sim difference, headroom correlates with the grounding benefit at
$r=\headroomR{}$ ($t=\headroomT{}$). Splitting at the median, grounding is worth
\headroomHiHalf{} PA on the high-headroom half and \headroomLoHalf{} on the low. Estimating headroom
from the Sim score itself would give $r=\headroomNaiveR{}$, but roughly half of that is regression
to the mean, since a task scoring low on Sim by chance will show a positive difference for that
reason alone. The weaker figure is the defensible one.

\subsection{Benchmark Saturation}
\label{sec:saturation}

The ceiling counts also bound what any comparison on this benchmark can show. On Claude, four of
six conditions place a majority of tasks at PA${}=1.000$: typed Sims (\ceilClaudeSim{}), grounded
Sims (\ceilClaudeSimG{}), flat profiles (\ceilClaudeFlat{}), and even retrieval reaches
\ceilClaudeRAG{}. The two comparisons that fail to reach significance on Claude, grounding and
structure-versus-flat, are the two whose conditions are both near ceiling.

This is a property of the instrument. \simbench{} discriminates well at
Llama~70B's capability level, weakly at GPT-4o's, and poorly at Claude's. Benchmarks of this kind
have a limited useful lifetime as models improve, and reporting the ceiling distribution alongside
mean scores is a cheap way to make that visible.

Table~\ref{tab:mde} makes the pattern explicit. The benchmark resolves \resolvedLlama{} of five
contrasts on Llama~3.3~70B, \resolvedGPT{} on GPT-4o and \resolvedClaude{} on Claude Sonnet~4.5:
its discriminating power falls monotonically as model capability rises. The unresolved cells are
not scattered at random but concentrate on the smallest contrasts for the strongest models, which
is what saturation predicts.

\section{Exploratory: Parametric Personalization}
\label{sec:lora}

This section is exploratory. Its metrics measure surface similarity to a reference, not
Preference Adherence, so its results do not bear on the claims above.

Prompt-level representations condition a model at inference time. A complementary question is
whether the same typing can guide weight-level adaptation, and at what granularity. Prior work
assigns one adapter per user~\citep{tan2024oppu}, which scales poorly and, at realistic per-user
data volumes, has little data to learn from. We test whether \emph{Sim type} is a better unit than
user identity.

\paragraph{Setup.}
We train LoRA adapters~\citep{hu2021lora} on \loraBase{}~\citep{jiang2023mistral} using Apple
MLX~\citep{apple2023mlx}, with rank
$r=\loraRank{}$, $\alpha=\loraAlpha{}$, learning rate $\loraLR{}$, batch size \loraBatch{},
\loraLayers{} adapted layers, maximum sequence length \loraSeqLen{}, trained for \loraIters{}
iterations. Five arms are compared, all evaluated on the same frozen set of \loraN{} held-out
examples with an identical prompt builder:

\begin{itemize}
\item \textbf{Base}: no adapter.
\item \textbf{Domain-mismatched}: one adapter trained on Amazon product reviews~\citep{ni2019amazon}, which are large,
      stylistically coherent, semantically distant from preference-conditioned planning.
\item \textbf{Task-matched}: one adapter trained on synthetic planning data in the target format.
\item \textbf{Per-user}: ten adapters, one per synthetic user, each trained on that user's
      data only (the OPPU setting).
\item \textbf{Sim-type routed}: \loraNumAdapters{} adapters, one per Sim type (work, family,
      health), selected at inference by a lightweight classifier over the task request.
\end{itemize}

\begin{table}[t]
\caption{Adapter arms on a frozen held-out set, $n=\loraN{}$. All arms share the same eval set and
prompt construction. Routing accuracy for the Sim-type arm is \loraRoutingAcc{}.}
\label{tab:lora}
\centering
\small
\begin{tabular}{lcc}
\toprule
Adapter & ROUGE-L & BERTScore \\
\midrule
Base (none)          & \loraBaseR{}      & \loraBaseB{} \\
Domain-mismatched    & \loraAmazonR{}    & \loraAmazonB{} \\
Task-matched         & \loraSyntheticR{} & \loraSyntheticB{} \\
Per-user ($\times$10)& \loraPerUserR{}   & \loraPerUserB{} \\
Sim-type routed ($\times$\loraNumAdapters{}) & \textbf{\loraRoutedR{}} & \textbf{\loraRoutedB{}} \\
\bottomrule
\end{tabular}
\end{table}

\paragraph{Granularity does more work than the adapter.}
Table~\ref{tab:lora} gives the five arms.
Both multi-adapter arms improve over base by a wide margin (per-user \loraPerUserVsBase{} and
Sim-type routed \loraRoutedVsBase{} ROUGE-L, both $p<0.001$), while on ROUGE-L the two single-adapter arms
barely move: domain-mismatched \loraAmazonVsBase{}, task-matched \loraSyntheticVsBase{}. BERTScore
tells a different story for the task-matched arm, which rises from \loraBaseB{} to
\loraSyntheticB{}. Training
on in-format data buys far less than conditioning on \emph{which} user context applies.

\paragraph{Sim type is a better unit than user identity.}
Routing \loraNumAdapters{} Sim-type adapters outperforms maintaining ten per-user adapters by
\loraRoutedVsPerUser{} ROUGE-L (95\% CI \loraRoutedVsPerUserCI{}, $p=\loraRoutedVsPerUserP{}$,
paired over examples). The effect is modest and we report its detection threshold alongside it:
the paired MDE at $n=\loraN{}$ is \loraMDE{}, so the estimate is resolved but not by a wide margin.
What makes it notable is the ratio: three adapters match and slightly exceed ten, with routing
accuracy of only \loraRoutingAcc{}, so the advantage survives frequent misrouting. Sim type
partitions users into groups that share enough behaviour to pool training data, which is the
scaling property per-user adaptation lacks.

\paragraph{What these numbers do not show.}
ROUGE-L~\citep{lin2004rouge} and BERTScore~\citep{zhang2020bertscore} measure surface similarity to a reference completion, not whether a plan
satisfies the user's constraints. They therefore speak to whether Sim-typed routing produces
text closer to the target, not to plan quality in the sense our main results measure it. Applying
Preference Adherence to adapted models requires generating \simbench{} plans from them and scoring
with the same judge, which we have not done; it is the natural next step and we flag the gap rather
than treating these metrics as a proxy for it.

\section{Limitations}
\label{sec:limitations}

\paragraph{Synthetic profiles.}
Our \nUsers{} users are authored, so their Sims are clean by construction: constraints typed,
priorities weighted, contexts separated. Real interaction logs are none of these things. Everything
below holds given a well-formed multi-context profile; extracting one from unstructured behavioural data is
the harder problem and we have not touched it.

\paragraph{What the benchmark cannot resolve.}
At $n=\nTasks{}$ the per-comparison detection threshold runs from \mdeMin{} to
\mdeMax{} PA. Five contrasts fall below their own threshold and we mark them unresolved:
grounding on GPT-4o (\groundingGPT{}) and Claude (\groundingClaude{}), typed structure against
flat concatenation on GPT-4o (\simVsFlatGPT{}) and Claude (\simVsFlatClaude{}), and removing
arbitration on Claude (\dClaudeSimA{}, exactly at its threshold). Detecting an effect
the size of the GPT-4o grounding estimate would take roughly \nNeededGPT{} tasks. We state the bound
because it cuts both ways: several format effects in this literature, including one we reported
ourselves in an earlier version of this work, sit below the resolution of the benchmarks used to
establish them. Saturation compounds the problem. Four of six conditions put a majority of Claude's
tasks at PA${}=1.000$, and a benchmark stops separating representations once a model solves most of
it. \simbench{} will lose its grip on frontier models well before it loses its grip on
mid-capability ones. The fix is harder tasks.

\paragraph{The judge is an imperfect instrument.}
Cross-judge agreement is moderate: per-plan Spearman \gridSpearman{}, Cohen's $\kappa$
\gridKappa{} on per-constraint verdicts, mean absolute difference \gridMAE{}, with our primary judge
scoring higher than the second by \gridOffset{}. The second judge reproduces the condition ranking
exactly, which is what our paired contrasts rely on, but it did not score the Claude planner's
plans.
A further \judgeLowWeightFrac{} of constraints get reduced weight from a lexical-overlap relevance
heuristic we never validated against human annotation. Our contrasts are within-model and paired at
the task level, so most of that bias differences out; absolute values and cross-model comparisons
carry it.

\paragraph{Grounding exemplars are derived, not observed.}
This one bothers us more than the others. The procedural examples are generated from the same Sim
definitions the model already sees, so in principle they cannot carry information the declarative
constraints lack, only a different presentation of it. That ceiling is a plausible mechanism for the
null on strong models: a model that can already operationalise a stated constraint gains little from
watching it be operationalised. A fair test needs exemplars drawn from real user decisions that
the profile fails to capture, which a synthetic setup cannot supply.

\paragraph{Generator drift.}
A same-day rerun of typed Sims on GPT-4o scored \simDrift{} below the July grid with identical
prompts and seeds. Every comparison in Table~\ref{tab:main} is within one run, and we keep
the Flat+Arb comparison inside its own same-day run, but absolute values from a served model
should not be compared across months.

\paragraph{Follow-up work.}
A companion study~\citep{shah2026counting} evaluates context selection on a revised and extended
benchmark, SimBench v1.2 (\url{https://github.com/VersarAI/SimBench}), scored against a fixed
human-annotated set of relevant constraints. That release removes the design notes described in the
\simbench{} section.

\section{Conclusion}

We set out to measure how much the format of a user representation affects
preference-conditioned planning, and which parts of \simguide{} carry the value. Three findings
stand out. Whether the agent receives structured user context at all
dominates every other choice. Among formats, \simguide{}'s typed multi-context blocks with
explicit arbitration beat retrieval
over the same user's past decisions by \simVsRagLlama{}, \simVsRagGPT{} and \simVsRagClaude{} PA on
Llama~3.3~70B, GPT-4o and Claude Sonnet~4.5, with the retrieval corpus generated from the same
underlying preferences so that the conditions differ in form rather than content. Most of that advantage comes from arbitration: removing it from typed blocks costs up to
\arbitrationLlamaAbs{} PA, while a flat profile given the same instruction does as well as typed blocks on the two
models where we could test it.

Procedural grounding, which we had previously reported as an unconditional benefit, is conditional
on the model having room to improve: \groundingLlama{} on the weakest model tested, falling to
\groundingGPT{} and \groundingClaude{} as baseline performance rises toward ceiling. We regard the
correction as the more useful result. It replaces a claim about a mechanism with a claim about when
the mechanism pays, and it identifies benchmark saturation as the variable that governs it.

For deployment the framework therefore reduces to a rule: always state how conflicts between
contexts are to be resolved, treat the typed block format as optional, and spend the extra prompt
length on grounding only where the model has headroom to use it.

Two methodological points generalise beyond this paper. Reporting a benchmark's minimum detectable
effect alongside its results costs one sentence and would have prevented us from publishing an
effect our own instrument could not resolve. And releasing a provenance audit that re-derives each
reported number from the prompt that produced it costs little and catches a class of error that
silently survives every conventional check, because well-formed prompts containing the wrong content
produce plausible results instead of failures. The benchmark, the evaluation code and the audit
are distributed with this preprint as ancillary files.

\end{document}